\documentclass{article}
\usepackage{float}

\usepackage[main, final]{neurips_2026}
\usepackage{amsmath}
\usepackage{tikz}
\usepackage{subcaption}
\usetikzlibrary{calc,positioning,fit,arrows.meta}
\usepackage{placeins}

\usepackage[utf8]{inputenc} 
\usepackage[T1]{fontenc}    
\usepackage{hyperref}       
\usepackage{url}            
\usepackage{booktabs}       
\usepackage{amsfonts}       
\usepackage{nicefrac}       
\usepackage{microtype}      
\usepackage{xcolor}         
\usepackage{subcaption}
\usepackage{graphicx} 

\usepackage{xspace}

\title{idSCD: Identifying Training Datasets through Semantic Correlation Descriptors}

\author{%
  Andrada Gobeajă \\
 POLITEHNICA University of Bucharest\\
  \texttt{andrada.gobeaja@stud.acs.upb.ro} \\
  \And
  Ionuț Hodoroagă \\
  POLITEHNICA University of Bucharest\\
  \texttt{ionut.hodoroaga@stud.acs.upb.ro} \\
  \And
  Marius Leordeanu \\
 POLITEHNICA University of Bucharest\\
 Institute of Mathematics of the Romanian Academy\\ 
  \texttt{marius.leordeanu@upb.ro} \\
    \And
  Elena Burceanu \\
  POLITEHNICA University of Bucharest\\
  \texttt{elena.burceanu@gmail.com} \\
}

\begin{document}

\maketitle

\begin{abstract}
Can a dataset be recognized from the spurious correlations it induces during training? We argue that datasets leave dataset-specific traces in a model’s learned semantic correlation structure: incidental regularities that are predictive within a dataset, but not causal for the underlying task, can be internalized during training. We use this insight to study dataset-level membership inference, moving beyond existing methods that rely on behavioral or distributional evidence such as confidence scores, losses, margins, generated samples, or query responses. We introduce a white-box semantic fingerprinting approach based on semantic correlation descriptors (SCDs), which capture the semantic correlation structure learned by a model and make it comparable across dataset mixtures. In a controlled leave-one-dataset-out diagnostic, SCDs recover dataset-specific changes and perfectly separate matching from non-matching dataset pairs. We then propose a practical SCD-based membership score that tests whether a target dataset is part of a model’s training mixture using only the model’s SCD and the target dataset’s standalone SCD, without requiring leave-one-dataset-out models.
Across three diverse experimental settings, with dataset groups for native language identification, emotion classification, and medical text classification, we test both the advantages and limitations of SCD-based membership inference with different degrees of semantic separation and keyword support between dataset splits. On average, the classifier \(id_{\mathrm{SCD}}\) based on this score achieves the highest performance and the lowest std, outperforming black-box baselines RMIA, Attack-P, and LiRA, as well as the white-box SIF baseline. These results show that dataset membership can be traced through internal semantic correlations, with the largest relative gain exceeding \(60\%\) in ROC-AUC when dataset groups expose distinct semantic particularities. Code publicly available  \href{https://github.com/andrada-gobeaja/idscd_semantic_correlation_descriptors}{HERE}.
\end{abstract}

\section{Introduction}
\label{sec:introduction}

Determining whether a dataset was used to train a model is increasingly important for model auditing. Modern models are trained on mixtures of public, private, licensed, synthetic, and domain-specific data, often with limited visibility into which sources were used. Dataset usage is therefore not merely a training detail: it affects privacy, licensing, benchmark contamination, reproducibility, and accountability. Prior work has studied related questions through membership inference, dataset-level membership inference, dataset inference, data provenance, and training-data attribution~\citep{carlini2022membershipinferenceattacksprinciples, maini2024llmdatasetinferencedid, sun2025enhancingtrainingdataattribution}.
Most existing evidence for dataset membership is output-driven. Methods typically rely on confidence scores, losses, prediction margins, generated samples, query responses, or influence estimates~\citep{carlini2022membershipinferenceattacksprinciples, ye2022enhancedmembershipinferenceattacks, cohen2022membershipinferenceattackusing}. These signals can be effective, but they primarily ask whether a model behaves as if it has seen a dataset. This makes them vulnerable to cases where different datasets induce similar output behavior, or where membership evidence is expressed less in predictions than in the internal structure learned by the model. We ask a complementary question: can a dataset be recognized from the semantic correlations it induces during training?

Our starting point is that datasets differ in incidental regularities arising from their construction, domain, annotation protocol, and label expression patterns. These regularities may be predictive within a dataset without being causal for the underlying task. When a model is trained on a mixture of datasets, such regularities can be internalized as spurious semantic correlations and persist in the trained model. Thus, even datasets mapped to the same label space may leave different fingerprints in how semantic features align with classes.
This suggests a different route to dataset-level membership inference: instead of probing the model only from the outside, we audit the semantic association structure learned inside the model. We introduce Semantic Correlation Descriptors (SCDs), which summarize this structure as dataset-level fingerprints and allow models trained on different mixtures to be compared in a shared correlation space. The goal is not to explain individual keywords, but to test whether the global semantic fingerprint of a target dataset is present in a model under inspection.

First, in a controlled leave-one-dataset-out diagnostic, SCDs recover dataset-specific changes and perfectly separate matching from non-matching dataset pairs, showing that the signal is not a generic training artifact. Second, in a more realistic membership setting, we use an SCD-based score that requires only the inspected model and a standalone reference descriptor for the target dataset, avoiding leave-one-dataset-out models. Across native language identification, emotion classification, and medical text classification, this classifier achieves the best average performance and lowest standard deviation against recent strong baselines, with the largest gains when dataset splits expose distinct semantic particularities. 

Our \textbf{main contributions} are the following:

\textbf{1) A semantic view of dataset-level membership inference.} Existing dataset-level membership methods primarily rely on behavioral or
    distributional evidence, such as confidence, loss, prediction margins, generated samples, or query responses. We instead formulate dataset membership
    as a semantic tracing problem: can we recover the presence of a target dataset from the high-level keyword-class correlations it induces in the trained model?
    
\textbf{2) Descriptor for dataset fingerprinting.}
    We introduce a Semantic Correlation Descriptor (SCDs), denoted \(\mathbf{d}_{\mathrm{SCD}}\),  a compact representation for
    keyword-class semantic correlation maps. Through a shared keyword vocabulary and
    zero-fill alignment, SCDs place models trained on different dataset mixtures in
    the same correlation space, enabling direct comparison of their learned
    association structure. In a controlled leave-one-dataset-out analysis, this
    representation perfectly separates matching from non-matching dataset pairs.
    
\textbf{3) A practical SCD-based membership score.}
    We propose a membership score that tests whether a target dataset \(D_i\) is part
    of the training mixture of a model under inspection, using only the model's SCD
    and the standalone SCD of \(D_i\). Unlike the controlled analysis, it does not
    require leave-one-dataset-out models. Across three diverse experimental
    settings for native language identification, emotion classification, and medical
    text classification, we test both the advantages and limitations of SCD-based
    membership inference with different degrees of semantic separation and keyword support between
    dataset splits. On average, the classifier \(id_{\mathrm{SCD}}\) induced by this
    score achieves the highest performance and the lowest standard deviation,
    outperforming recent strong baselines: the black-box methods RMIA, Attack-P,
    and LiRA, and the white-box SIF baseline.

\section{Related work}

\noindent\textbf{Relation to Spurious Correlations.} Spurious correlations arise when models rely on superficial, non-causal cues, such as backgrounds, textures, or secondary objects in vision tasks \citep{DBLP:journals/corr/abs-2004-07780, wang2021causalattentionunbiasedvisual,yang2022understandingrarespuriouscorrelations,lin2024shortcut} and keywords in language tasks \citep{gururangan2018annotationartifactsnaturallanguage,mccoy2019rightwrongreasonsdiagnosing,niven2019probingneuralnetworkcomprehension,du2023shortcutlearninglargelanguage}, instead of causal concepts. These shortcuts encode dataset-specific patterns that generalize poorly beyond the training distribution \citep{DBLP:journals/corr/abs-2004-07780, arjovsky2020invariantriskminimization}, often yielding higher confidence on samples containing familiar cues and greater uncertainty on out-of-distribution or clean test samples \citep{izmailov2022featurelearningpresencespurious,ye2024clever}. This confidence gap can be exploited by membership inference attacks \citep{nasr2018machinelearningmembershipprivacy, choquettechoo2021labelonlymembershipinferenceattacks}. We show that such shortcut patterns can also be recovered at the dataset level as semantic correlation fingerprints, enabling dataset-level membership inference.
 
\noindent\textbf{Relation to Membership Inference Attacks.}
Membership inference attacks (MIAs) determine whether a sample was used to train a machine learning model \citep{shokri2017membershipinferenceattacksmachine,carlini2022membershipinferenceattacksprinciples}. They exploit systematic differences between model behavior on training and unseen examples, and are typically classified by attacker access. White-box MIAs use internal signals such as activations, gradients, losses, confidence scores, parameters, or updates \citep{8406613,song2020systematicevaluationprivacyrisks,Rigaki_2023}, including gradient/activation attacks \citep{Nasr_2019}, Bayes-optimal loss tests \citep{sablayrolles2019whiteboxvsblackboxbayes}, memorized feature associations \citep{leino2020stolenmemoriesleveragingmodel}, and self-influence scores \citep{cohen2022membershipinferenceattackusing}. Black-box MIAs rely only on query outputs such as labels, confidence scores, logits, entropy, or losses \citep{li2021membershipleakagelabelonlyexposures,liu2022membershipinferenceattacksexploiting,carlini2022membershipinferenceattacksprinciples,zarifzadeh2024lowcosthighpowermembershipinference}, covering shadow-model attacks \citep{shokri2017membershipinferenceattacksmachine}, threshold attacks \citep{yeom2018privacyriskmachinelearning,salem2018mlleaksmodeldataindependent}, label-only attacks \citep{choquettechoo2021labelonlymembershipinferenceattacks,peng2024oslooneshotlabelonlymembership}, and calibrated likelihood-ratio attacks \citep{carlini2022membershipinferenceattacksprinciples,bertran2023scalablemembershipinferenceattacks}. Prior MIAs decide membership for individual examples using per-example signals. In contrast, we study dataset-level membership: given a target dataset $D_i$, we ask whether $D_i$ appeared during training, using the semantic correlation structure induced by the dataset as membership signal.

\noindent\textbf{Relation to Training Data Attribution.}
Training data attribution methods identify training examples that most affect model outputs using influence functions to approximate upweighting or removal effects  \citep{koh2020understandingblackboxpredictionsinfluence}, datamodels to predict counterfactual outputs from subset membership \citep{ilyas2022datamodelspredictingpredictionstraining}, and scalable variants such as TRAK \citep{park2023trakattributingmodelbehavior}, In-Run Data Shapley \citep{wang2025datashapleytrainingrun}, TrackStar \citep{chang2024scalableinfluencefacttracing}, and AirRep \citep{sun2025enhancingtrainingdataattribution}. These reduce attribution costs through kernel approximations, single-run contribution estimates, LLM-scale gradient influence, or learned attribution representations \citep{chang2024scalableinfluencefacttracing}.Our method targets dataset-level membership inference, asking whether an entire dataset was part of a model’s training mixture, using Semantic Correlation Descriptors (SCDs) built from keyword-class correlation maps to capture dataset-induced spurious correlations as dataset-specific signatures.

\begin{figure}[t]
  \centering
  \includegraphics[width=\linewidth]{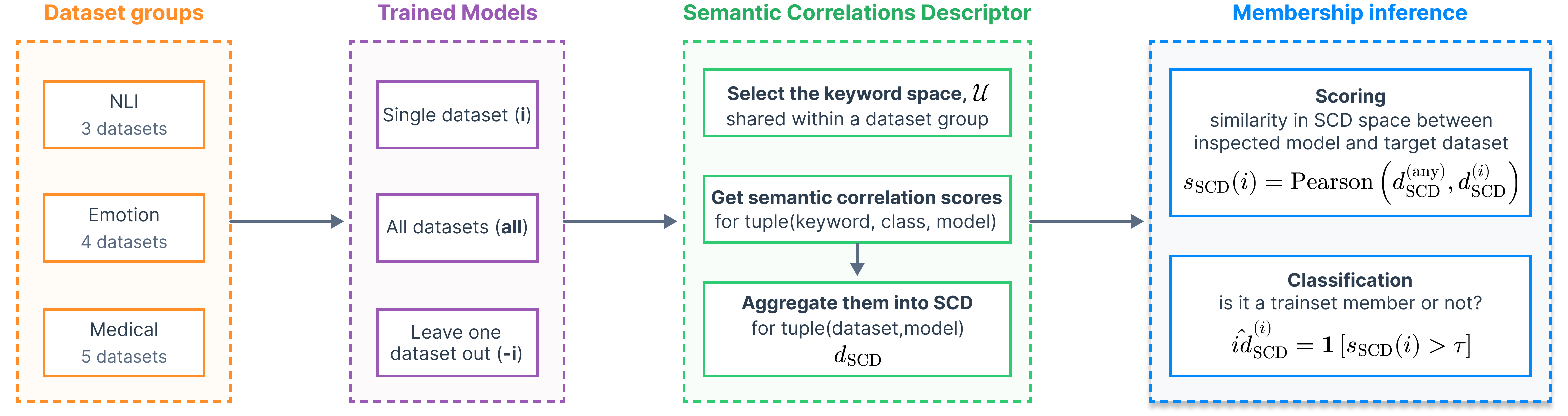}
    \caption{Dataset membership through semantic correlation fingerprints. The pipeline evaluates whether dataset-specific regularities leave recoverable traces in the semantic correlation structure of trained models. We extract semantic correlations over a keyword vocabulary shared within each dataset group, aggregate them into Semantic Correlation Descriptors (SCDs), and use the SCD classifier over the membership score to test whether a target dataset was part of the training mixture.}
    \label{fig:scd_pipeline}
\end{figure}

\section{Our approach}
\label{sec:approach}
Datasets differ in the incidental regularities that arise from their specific construction, domain, annotation protocol, and label expression patterns. During training, models may internalize these regularities as spurious correlations: surface keyword–class associations that are predictive within a dataset but noncausal for the underlying task.  In models trained on dataset mixtures, these correlations may become embedded in the learned decision structure and persist after training. We ask whether the presence of a target dataset can be recovered from the spurious correlations it induces during training. We study this question in a controlled setting with a group of datasets \(D_1,\ldots,D_N\), each mapped to a shared label space \(\mathcal{Y}\), and a fixed model architecture. Given a
target dataset \(D_i\), our goal is to infer whether \(D_i\) was present in the training mixture. 
To formalize this, let \(\mathcal{U}\) denote the global keyword vocabulary associated with the dataset collection. For each trained model \(m\), we construct a keyword-class correlation matrix:
\[
\mathbf{C}^{(m)} \in \mathbb{R}^{|\mathcal{U}| \times |\mathcal{Y}|},
\]
where each entry \(\mathbf{C}^{(m)}_{u,y}\) quantifies the strength of the correlation between keyword \(u \in \mathcal{U}\) and label
\(y \in \mathcal{Y}\) in model \(m\). Rather than interpreting individual keyword-class correlations, we treat the
full correlation map as a dataset-level fingerprint and use it to infer whether \(D_i\) was present in the training mixture through its dataset-specific spurious correlation patterns.

\subsection{From semantic correlations to dataset fingerprints}
\label{sec:fingerprints_contributions}
For each dataset group, we train three model variants designed to make dataset-specific contributions observable. For each target dataset \(D_i\), we first train a single-dataset model \(m_i\) using only \(D_i\). We also train a full model \(m_{\mathrm{all}}\) on the union of all datasets, \(D_{\mathrm{all}} = \bigcup_{i=1}^{N} D_i\), and a leave-one-dataset-out model \(m_{-i}\) on the union of all datasets except \(D_i\), \(D_{-i} = \bigcup_{j \neq i} D_j\).

\noindent\textbf{Semantic correlation maps.}
Our approach requires a scoring function that estimates the degree of
keyword-class correlation across trained models. In our implementation, we
obtain the correlation matrices \(\mathbf{C}\) using BEE~\citep{paduraru2026bee},
which extracts keyword-class correlations by comparing keyword embeddings and
class weight vectors in a shared embedding space.
In BEE, each trained model is converted into a keyword-class correlation matrix.
For a model \(m\), each keyword \(u \in \mathcal{U}\) is embedded with the
model's last-token hidden state, giving a keyword embedding
\(\mathbf{e}_u^{(m)}\). Let \(\mathbf{w}_y^{(m)}\) be the classifier's
vector for class \(y\), extracted after training, from the sequence-classification head. The
keyword-class alignment is computed by cosine similarity,
\(\mathrm{sim}_m(u,y)=\cos\!\left(\mathbf{e}_u^{(m)}, \mathbf{w}_y^{(m)}\right)\).
The cosine alignment is then refined into the BEE scoring function for
keyword-class correlations - which yields entries of the correlation matrix
where rows correspond to keywords and columns to labels:
\begin{equation}
\mathbf{C}^{(m)}_{u,y} = \mathrm{BEE}_m(u,y)
=
\mathrm{sim}_m(u,y)
-
\min_{y'\in\mathcal{Y}}\mathrm{sim}_m(u, y').
\end{equation}

\noindent\textbf{Zero-fill alignment.}
Keyword extraction tools generally produce keyword sets of different sizes across datasets. As a result, the number of rows of the corresponding correlation matrices \(\mathbf{C}^{(m_i)}\) may differ. To bring these matrices to the same size we use a shared global keyword vocabulary \(\mathcal{U}\) within each dataset group. 
For each model \(m_i\), we construct \(\mathbf{C}^{(m_i)}\) over the same vocabulary \(\mathcal{U}\). If keyword \(u\) is present in the keyword set
extracted for model \(m_i\), its BEE scores are inserted into the corresponding
row of \(\mathbf{C}^{(m_i)}\). If \(u\) is absent, that row is set to zero: 

\begin{equation}
\mathbf{C}^{(m_i)}_{u,y}
=
\begin{cases}
\mathrm{BEE}_{m_i}(u,y), & \text{if keyword } u \text{ is present for model } m_i,\\[4pt]
0, & \text{otherwise.}
\end{cases}
\end{equation}

This zero-fill alignment yields correlation matrices with a common shape,
\(|\mathcal{U}| \times |\mathcal{Y}|\), enabling direct comparison of trained models from the perspective of their learned keyword-class correlations.

\noindent\textbf{Semantic Correlation Descriptor (SCD).}
For each dataset \(D_i\), we define its correlation descriptor as the vectorized correlation map of the corresponding single-dataset model \(m_i\): $d_{\mathrm{SCD}}^{(i)} = \mathrm{vec}\!\left(\mathbf{C}^{(m_i)}\right)$
By construction, \(d_{\mathrm{SCD}}^{(i)}\) captures the keyword-class
correlation pattern learned from \(D_i\) in isolation. This pattern includes the spurious correlations induced by the dataset, and serves as a dataset-specific reference for comparing learned correlation structures.

\subsection{SCD sensitivity analysis in a controlled setting}
\label{sec:cd_sensitivity}

Before defining the final membership score, we first test whether SCDs capture
dataset-specific correlation patterns in a controlled setting. For each target
dataset \(D_i\), we compare the full model \(m_{\mathrm{all}}\) with the
leave-one-dataset-out model \(m_{-i}\). The goal of this analysis is not to
define the final membership inference protocol, but to verify whether the
change between these two models aligns with the standalone descriptor of
\(D_i\), rather than with the descriptors of other datasets.
We estimate the dataset-specific change associated with \(D_i\) as
\begin{equation}
\Delta_i =
d_{\mathrm{SCD}}^{(\mathrm{all})}
-
d_{\mathrm{SCD}}^{(-i)} .
\label{eq:delta}
\end{equation}
Intuitively, \(\Delta_i\) captures the change in the learned keyword-class
correlation structure associated with \(D_i\). If \(D_i\) induces a distinctive
correlation pattern, then this estimated change should be more similar to the
standalone descriptor of \(D_i\) than to the descriptors of other datasets.
We measure this alignment using Pearson correlation. For each target dataset
\(D_i\) and each candidate dataset \(D_j\), we define
\begin{equation}
s_{\Delta}(i,j)
=
\mathrm{Pearson}
\left(
\Delta_i,
d_{\mathrm{SCD}}^{(j)}
\right)
=
\mathrm{Pearson}
\left(
d_{\mathrm{SCD}}^{(\mathrm{all})}
-
d_{\mathrm{SCD}}^{(-i)},
d_{\mathrm{SCD}}^{(j)}
\right).
\label{eq:attrib_score_delta}
\end{equation}

This score induces a binary test over dataset pairs \((i,j)\): pairs with
\(j=i\) correspond to matching descriptors, while pairs with \(j \neq i\)
correspond to non-matching descriptors. A simple threshold classifier can
therefore be defined as $\hat{z}_{i,j} = \mathbf{1} \left[s_{\Delta}(i,j) > \tau\right]$,
where \(\hat{z}_{i,j}=1\) predicts that \(\Delta_i\) matches the descriptor of
\(D_j\), and \(\hat{z}_{i,j}=0\) otherwise.

The resulting matrix \(S_{\Delta} \in \mathbb{R}^{N \times N}\) measures how
well each estimated change matches each standalone dataset descriptor. Diagonal dominance, \(s_{\Delta}(i,i) > s_{\Delta}(i,j)\) for all
\(j \neq i\), indicates that the change associated with \(D_i\) is best
explained by the fingerprint of \(D_i\) itself. In Sec.~\ref{sec:experiments},
we use this controlled analysis to validate the dataset-specificity of SCDs: the
threshold classifier based on \(s_{\Delta}\) perfectly separates matching from
non-matching pairs, achieving a ROC-AUC of \(100\%\) in
Fig.~\ref{fig:cd_sensitivity_heatmaps}. This controlled analysis is not intended as the final membership inference
protocol, since it assumes access to both \(m_{\mathrm{all}}\) and
\(m_{-i}\). Rather, it serves as a diagnostic test of whether SCDs respond to
dataset-specific changes in the learned correlation structure. In the next
section, we consider a more realistic setting and define a membership score
that does not require leave-one-dataset-out models.

\subsection{SCD-based dataset membership inference}
\label{sec:attribution_membership}
 
Instead of
requiring paired models trained with and without a target dataset \(D_i\), we
assume access only to a trained model \(m_{\mathrm{any}}\) and to the
single-dataset reference descriptor of \(D_i\). The goal is to decide whether
\(D_i\) was included in the training data of \(m_{\mathrm{any}}\).

The key idea is that, if \(m_{\mathrm{any}}\) was trained on \(D_i\), then the
dataset-specific correlation pattern captured by the descriptor of \(D_i\)
should also be visible in the correlation descriptor of \(m_{\mathrm{any}}\).
We therefore compare the two descriptors directly using Pearson correlation:
\begin{equation}
s_{\mathrm{SCD}}(i)
=
\mathrm{Pearson}
\left(
d_{\mathrm{SCD}}^{(\mathrm{any})},
d_{\mathrm{SCD}}^{(i)}
\right).
\end{equation}

Higher values of \(s_{\mathrm{SCD}}(i)\) indicate stronger evidence that the
correlation pattern associated with \(D_i\) is present in \(m_{\mathrm{any}}\).
This score essentially measures how strongly
the descriptor of the target dataset matches the descriptor of the model under
inspection. Unlike \(s_{\Delta}\), it does not require training or accessing
models explicitly constructed with \(D_i\) and without \(D_i\), 
making it suitable for real-life membership inference settings. We use \(s_{\mathrm{SCD}}\) to define a threshold-based membership classifier:
\begin{equation}
\hat{id}_{\mathrm{SCD}}^{(i)}
=
\mathbf{1}
\left[
s_{\mathrm{SCD}}(i) > \tau
\right],
\label{eq:idscd}
\end{equation}
where \(\hat{id}_{\mathrm{SCD}}^{(i)}=1\) predicts that \(D_i\) is included in the training data of
\(m_{\mathrm{any}}\), and \(\hat{id}_{\mathrm{SCD}}^{(i)}=0\) otherwise. In
Sec.~\ref{sec:experiments}, we evaluate this classifier by measuring the
ROC-AUC obtained from \(s_{\mathrm{SCD}}\) in Fig~\ref{fig:sota_comparison}.

\section{Experiments}
\label{sec:experiments}
 
\noindent\textbf{Dataset groups.} We evaluate dataset-level attribution and membership inference on three dataset groups: native language identification (NLI), emotion classification, and medical section classification. The emotion group contains four heterogeneous datasets: EmoLit \citep{app13137502}, EmotionDataset20 \citep{emotion_dataset_20_2025}, GoEmotions \citep{demszky2020goemotions}, and a merged XED \citep{ohman2020xed} /SMED \citep{ganguly_smed_kaggle} corpus. The NLI group contains three datasets of English text written by non-native speakers: Reddit-L2, EFCAMDAT, and iTalki. The medical section classification group is derived from the PubMed 200k RCT corpus \citep{Dernoncourt2017PubMed2R}. In every group, all datasets are mapped to a shared set of common classes so that keyword-class association maps are comparable across models trained on different dataset subsets. 
 
\noindent\textbf{Emotion dataset group.}
The emotion group combines four sources that differ in genre, granularity, and annotation protocol. \textbf{EmoLit} is treated as multi-label over fine-grained emotion dimensions: after deduplicating sentences, we assign each example the emotion dimension with the largest real-valued score and then map that winner to a coarse label using a fixed many-to-one scheme. \textbf{GoEmotions} merges the train, development, and test splits; we drop duplicate comments, and for multi-labeled rows we take the \emph{first} emotion id in the comma-separated list as the provisional fine-grained tag before mapping to the same coarse inventory. \textbf{EmotionDataset20} uses the CSV text field and single per-sentence emotion string; labels are normalized (lowercased) and mapped with the same coarse taxonomy. \textbf{XED} (English) supplies integer ids that may enumerate several emotions per line. We expand these into one row per id, map ids to string emotion names, and concatenate the result with \textbf{SMED}, after normalizing SMED's string labels (strip, title case). The XED/SMED union is deduplicated on raw text. All four mapped corpora are concatenated and deduplicated again on \texttt{text} (first occurrence kept) to form a single pooled emotion dataset, while leave-one-corpus-out training files are built by concatenating any three of the four mapped sources with the same text-level deduplication (see App.~\ref{app:dataset_configurations} for details). Throughout, fine-grained original labels are collapsed into nine shared classes: \emph{Angry}, \emph{Anticipation}, \emph{Disgust}, \emph{Fear}, \emph{Happy}, \emph{Neutral}, \emph{Sad}, \emph{Surprise}, and \emph{Trust}, so that keyword-class association rules remain comparable across models trained on different subsets.

\noindent\textbf{NLI datasets.} The NLI evaluation relies on three foundational datasets: Reddit-L2 \citep{rabinovich2018nativelanguagecognateeffects}, EFCAMDAT \citep{geertzen2014efcamdat}, and iTalki \citep{hudson2018development}. To avoid trivial membership inference signals caused by unique labels, we filter the corpora to retain only the five first-language (L1) classes shared by all three sources: French, German, Italian, Spanish, and Turkish. To ensure the models are not biased toward the larger corpora, we use stratified subsampling to Reddit-L2 and EFCAMDAT, reducing them to the size of the smallest dataset (iTalki), as detailed in App.~\ref{app:dataset_configurations}, while preserving original class distributions. Due to its prohibitive size, Reddit-L2 is subsampled by a chunked, two-pass approach and EFCAMDAT is processed in-memory.
 
\noindent\textbf{Medical datasets.} Unlike the emotion and NLI groups, which inherently consist of multiple distinct datasets, the medical dataset group is a five-way split created from PubMed 200k RCT, a single large biomedical corpus containing over 2 million sentences. Following prior work on clustering-based splits~\citep{firca2025notallsplits}, we partition it into semantically distinct topic clusters expected to induce clearer dataset fingerprints. First, to accommodate the computational footprint of the 14-million parameter LLM, Pythia \citep{biderman2023pythiasuiteanalyzinglarge}, used in our experiments, we extract a 12\% stratified subsample of the original dataset. This subsample preserves the original class distribution of the rhetorical roles (i.e., BACKGROUND, METHODS, RESULTS, CONCLUSIONS, OBJECTIVE). To ensure the data is split based on overarching medical topics rather than sentence-level syntax, we aggregate the individual sentences using their \texttt{abstract\_id} to reconstruct the full text of each abstract. We apply TF-IDF \citep{sparckjones1972statistical} (Term Frequency-Inverse Document Frequency) to these reconstructed abstracts to extract features, removing standard English stop words and limiting the vocabulary to the top 10,000 terms. We then apply the MiniBatch K-Means \citep{sculley2010web} algorithm to the resulting TF-IDF matrix to group the abstracts into five distinct topic clusters. Crucially, we map these cluster assignments back to the original sentence-level data. This ensures that all sentences belonging to a single abstract are kept together in the same cluster, strictly preventing data leakage across different splits.

\noindent\textbf{Keyword extraction.}
Keywords are extracted independently for each trained-model configuration using YAKE~\citep{campos2020yake}. We use YAKE with maximum n-gram length \(n=3\), \texttt{dedupLim=0.9}, and 5000 keywords per model configuration, the resulting keyword pool containing unigrams, bigrams, and trigrams. We filter class-name keywords and empty or invalid tokens. For each dataset group, we construct a global keyword pool \(\mathcal{U}\) as the union of all extracted keywords across the full, leave-one-out, and single-dataset models in that group.

\noindent\textbf{Implementation details.}
We use the Pythia-14M architecture (\texttt{EleutherAI/pythia-14m}) as the base model. Each model is first finetuned as a causal language model and then used to initialize a sequence-classification model. The classifier uses the \texttt{GPTNeoXForSequenceClassification} head, exposed as \texttt{model.score}; the class directions used by BEE are the rows of \texttt{model.score.weight}. Both stages are trained for three epochs with learning rate \(2\times10^{-5}\), weight decay 0.01, epoch-level evaluation and checkpointing, and mixed precision when CUDA is available. 
All preprocessing, sampling, training, and evaluation use fixed random seed $42$. For each dataset group, we train individual, leave-one-dataset-out, and full-corpus models, generating 27 training configurations in total, as reported in App.~\ref{app:dataset_configurations}.

\noindent\textbf{Deployment assumptions.} Deploying idSCD assumes: 
\textit{(i) Model access}: white-box access to the inspected model, specifically its last-layer embeddings and classification-head weights, required to construct its SCD. In our implementation, these correspond to the last-token hidden state $\mathbf{e}_u^{(m)}$ for each keyword $u$ and the class directions $\mathbf{w}_y^{(m)}$ from the sequence-classification head. \textit{(ii) Target reference}: access to the target dataset to construct its standalone SCD by training a reference model from the same pretrained checkpoint as the inspected model, using the same SCD construction procedure. The deployment score $s_{\mathrm{SCD}}$ directly compares the inspected and target-reference SCDs, without requiring a leave-one-dataset-out version of the inspected model.
\textit{(iii) Shared label space}: candidate datasets must share, or be mapped to, a common label space, making their SCD similarity scores directly comparable and enabling calibration and interpretation of the target dataset’s membership score.
\textit{(iv) Semantic support}: candidate datasets must provide sufficient shared keyword support, as low overlap reduces the semantic evidence and may weaken the membership signal. Keyword overlap can be measured a priori, before model training, from the raw datasets using the keyword extraction procedure (see App.~\ref{app:keyword-overlap}).

\noindent\textbf{Computational costs.} Model training was run on NVIDIA A100 GPUs with 40GB VRAM on Google Colab, while dataset processing and keyword extraction used high-RAM CPU workers with 40GB RAM. Reproducing the main experiments requires approximately 42 GPU-hours: 19 hours for the emotion group, 9 hours for NLI, and 14 hours for the medical group. Reproducing the baseline comparisons requires an additional 30 GPU-hours.
 
\noindent\textbf{Ablations and design choices.}
We explored several variants along the pipeline before settling on the final setup. At the preprocessing level, we varied YAKE keyword extraction by changing the \(n\)-gram length (\(n=\overline{1,5}\)), the number of extracted keywords per dataset (\(200,500,1000,2000,5000\)), and class-conditioned extraction strategies such as selecting the top \(\overline{10, 100}\) keywords per class. We also tested alternative medical split constructions, including PCA followed by K-Means, before adopting TF-IDF with K-Means, and initially treated XED and SMED as separate emotion sources before merging them into a single dataset. At the descriptor construction level, we evaluated variants without zero-fill alignment, variants that did not merge matrices into a shared vocabulary space, and alternative BEE scoring choices, including softmax-temperature variants with \(T=\overline{1,3}\). At the scoring level, we tested cosine similarity instead of Pearson correlation, as well as direct Pearson correlation over logits rather than BEE-based descriptors. These ablations guided the final design: shared-vocabulary zero-fill alignment, BEE-based semantic correlation descriptors, and Pearson-based SCD scoring.

\begin{figure}[t]
    \centering
    \begin{subfigure}[t]{0.27\textwidth}
        \centering
        \includegraphics[width=\linewidth]{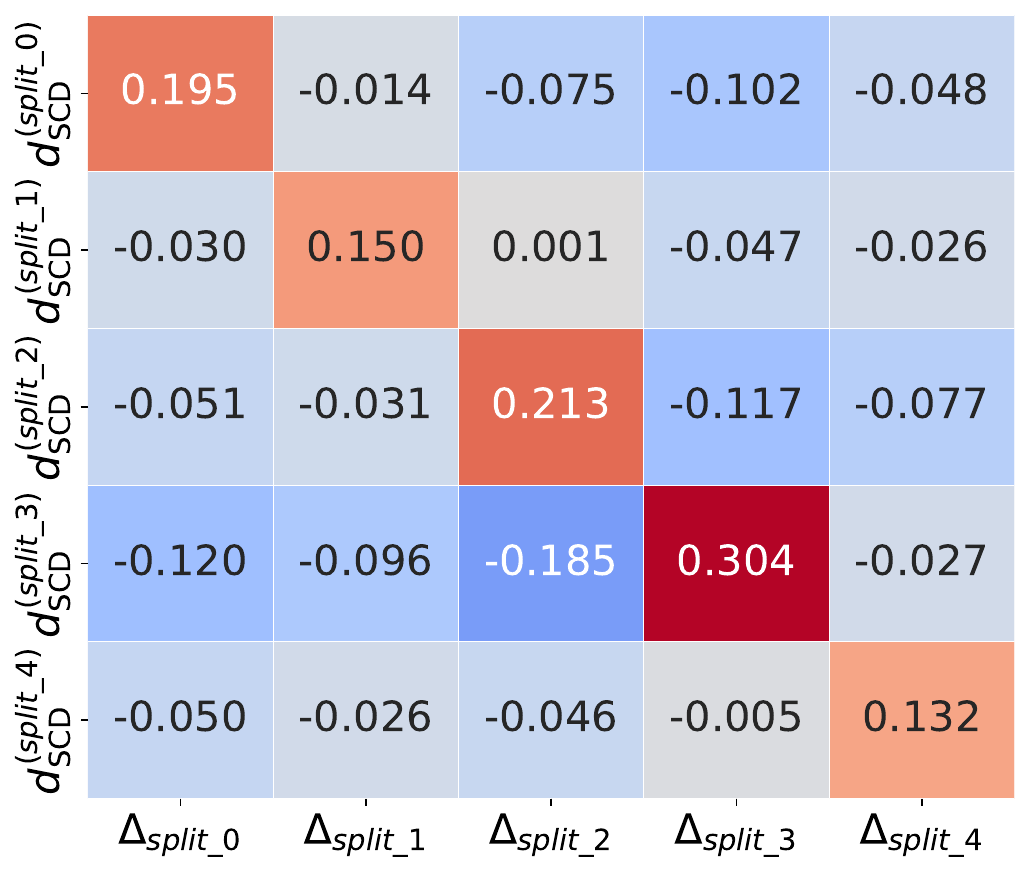}
        \caption{Medical datasets group.}
        \label{fig:datasets_pubmed}
    \end{subfigure}
    \hfill
    \begin{subfigure}[t]{0.27\textwidth}
        \centering
        \includegraphics[width=\linewidth]{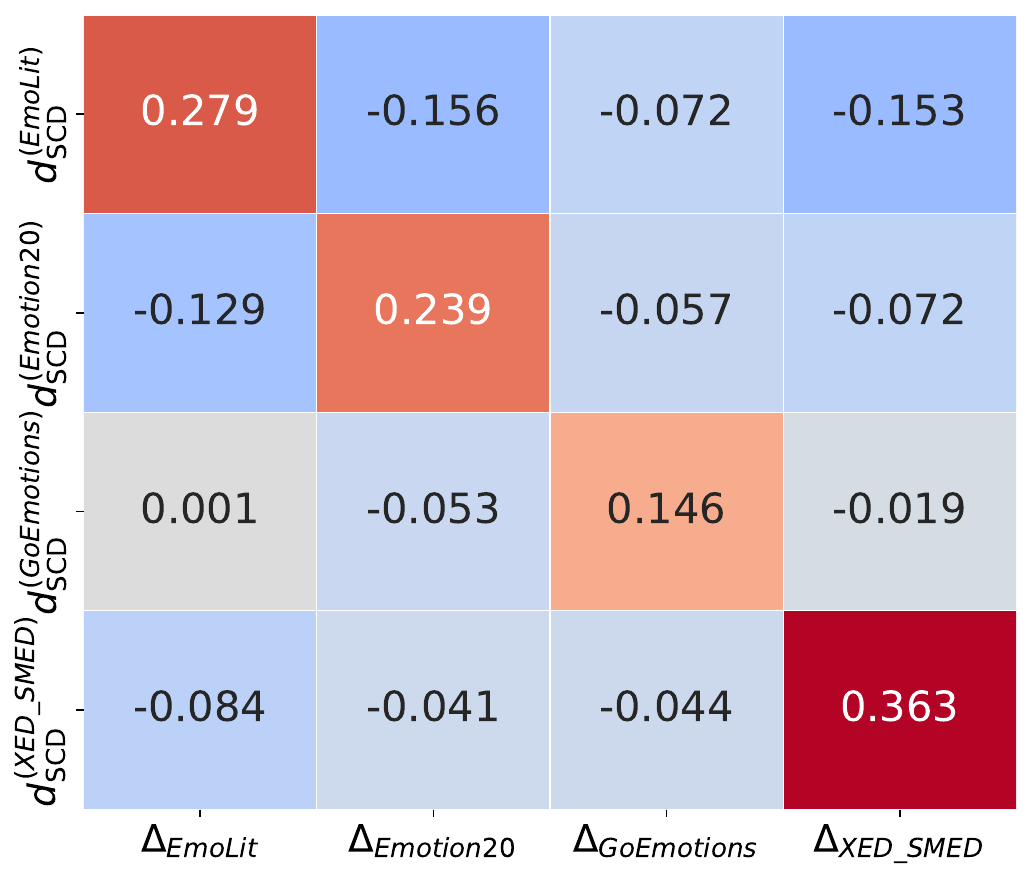}
        \caption{Emotion datasets group.}
        \label{fig:datasets_emotion}
    \end{subfigure}
    \hfill
    \begin{subfigure}[t]{0.27\textwidth}
        \centering
        \includegraphics[width=\linewidth]{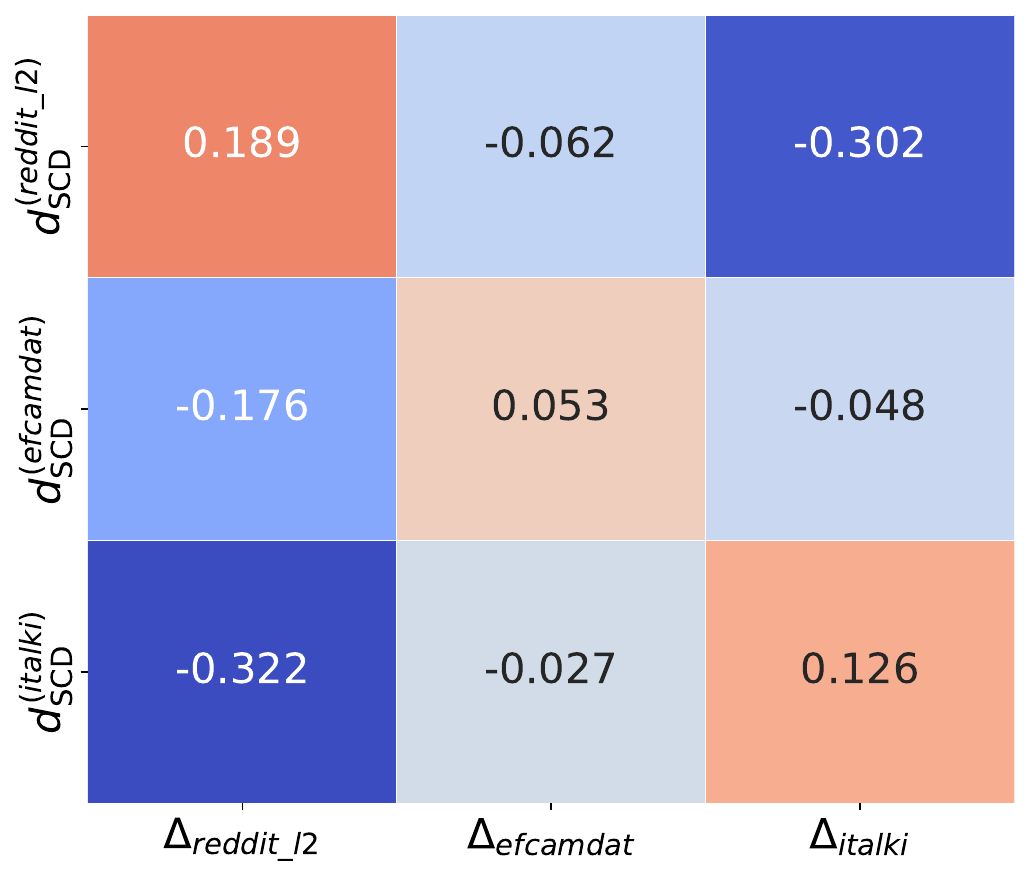}
        \caption{NLI datasets group.}
        \label{fig:datasets_nli}
    \end{subfigure}
    \caption{
        Dataset-specificity of correlation descriptors in the controlled
        difference-based setting. Each heatmap shows the \(S_{\Delta}\) matrix defined
        in Sec.~\ref{sec:cd_sensitivity}, with rows corresponding to target datasets
        and columns corresponding to standalone dataset descriptors, defined in Eq. ~\ref{eq:delta}. Across different dataset groups, the strongest score in every row lies on
        the diagonal, showing that each estimated dataset-specific change is best
        matched by the descriptor of the corresponding dataset. Treating diagonal
        entries as positives and off-diagonal entries as negatives yields perfect
        separation, with a ROC-AUC of \(100\%\).}
        \label{fig:cd_sensitivity_heatmaps}
\end{figure}

\subsection{Correlation Descriptors are dataset-specific}
\label{sec:exp_cd_sensitivity}

We evaluate here whether the correlation descriptors defined in
Sec.~\ref{sec:fingerprints_contributions} are sensitive to dataset-specific
changes in the training mixture. This experiment is designed as a controlled
diagnostic for the intuition in Sec.~\ref{sec:cd_sensitivity}: if a dataset
\(D_i\) contributes a characteristic correlation pattern, then the
difference-based descriptor \(\Delta_i\) from Eq.~\ref{eq:delta} should align
most strongly with the standalone descriptor of \(D_i\), rather than with the
descriptors of other datasets.
 
\noindent\textbf{Results.} Fig.~\ref{fig:cd_sensitivity_heatmaps} shows the resulting
\(S_{\Delta}\) matrices for the NLI, emotion, and medical dataset groups. The
pattern is consistent across all groups: the largest score in each row appears
on the diagonal. This means that the estimated change associated with each
target dataset is best matched by that dataset's own correlation descriptor.
Off-diagonal entries are substantially lower and often negative, indicating
that these changes do not align with unrelated dataset descriptors. We further quantify this separation by treating diagonal entries
\((i=j)\) as positives and off-diagonal entries \((i\neq j)\) as negatives. A
simple threshold classifier over \(s_{\Delta}(i,j)\) perfectly separates the two
classes, achieving a ROC-AUC of \(100\%\). This confirms that, in this
controlled setting, correlation descriptors are not merely capturing generic
training artifacts. They encode dataset-specific correlation structure strongly
enough to recover which dataset accounts for the observed change in the model.  The experiments were repeated with five random seeds,  and the results are highly stable across all dataset groups as seen in \ref{app:multi-seed-experiments}
 
\noindent\textbf{Takeaway.}
This experiment validates the core mechanism behind our approach:
dataset-specific correlation patterns are measurable, separable, and
recoverable from trained models. At the same time, this is a controlled
diagnostic setting because it assumes access to the leave-one-dataset-out model
\(m_{-i}\). In the next section, we move to a more realistic membership
inference setting, where we do not rely on leave-one-dataset-out models.

\subsection{SOTA Comparison for Membership Inference}
\label{sec:exp_sota_comparison}

 We next compare the practical SCD-based membership classifier from Sec.~\ref{sec:attribution_membership} against established membership inference baselines. Unlike the controlled diagnostic, this setting does not rely on leave-one-dataset-out differences. The goal of this evaluation is to test whether \(id_{\mathrm{SCD}}\) can distinguish models that were trained with a target dataset from models where that dataset was absent. For each target dataset $D_i$, member observations are computed with the full model $m_{all}$ and non-member observations with the leave-one-dataset-out model $m_{-i}$. In both cases, we obtain one dataset-level observation by correlation the full per-sample score vectors, and one class-level observation for each label $y \in Y_i$ by restricting the score vectors to the samples of class y. Each target dataset therefore contributes $1 + |Y_i|$ member observations and $1 + |Y_i|$ non-member observations. We compute ROC-AUC by varying the decision threshold of \(id_{\mathrm{SCD}}\), and compare against recent strong membership inference baselines: black-box methods RMIA, Attack-P, and LiRA, which rely on output-derived membership signals such as loss, confidence and likelihood-ratio, and the white-box SIF baseline, which uses model-internal information to compute self-influence scores. For a fair comparison, all baselines use the same dataset-level fingerprint protocol, with each model’s per-sample membership score vector serving as its fingerprint $s(m)$. For each target dataset $D_i$, we compute Pearson correlations over both the full vectors and each class-restricted subset. Member and non-member observations are $\operatorname{Pearson}(s(m_{all}), s(m_i))$ and $\operatorname{Pearson}(s(m_{-i}), s(m_i))$, respectively, where $m_{all}$, $m_i$, and $m_{-i}$ denote the full, single-dataset, and corresponding leave-one-dataset-out models.
 
\begin{figure}[t]
    \centering
        \begin{subfigure}[t]{0.24\linewidth}
        \centering        \includegraphics[width=\linewidth]{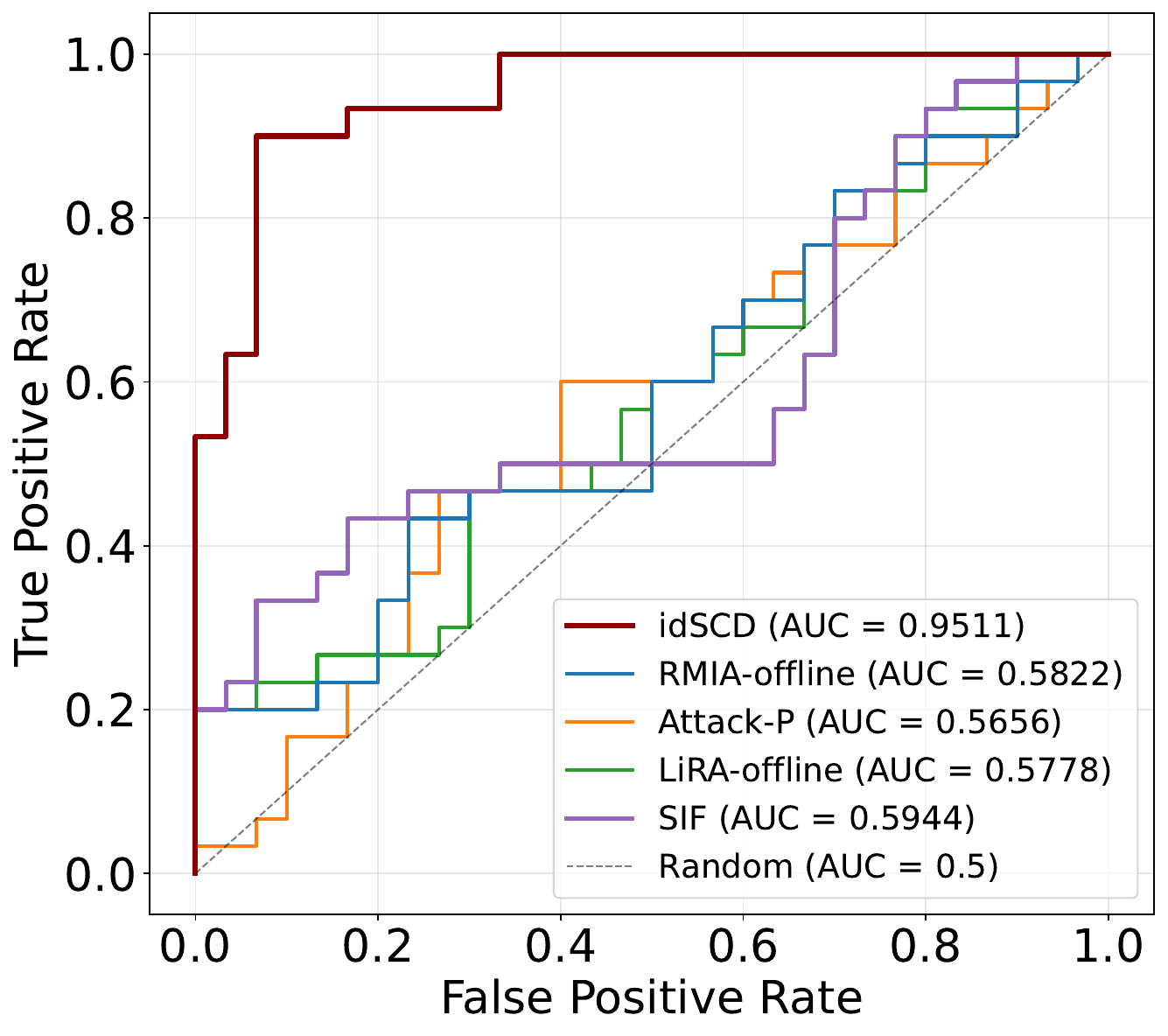}
        \caption{Medical datasets group.}
        \label{fig:sota_medical}
    \end{subfigure}
    \begin{subfigure}[t]{0.24\linewidth}
        \centering
        \includegraphics[width=\linewidth]{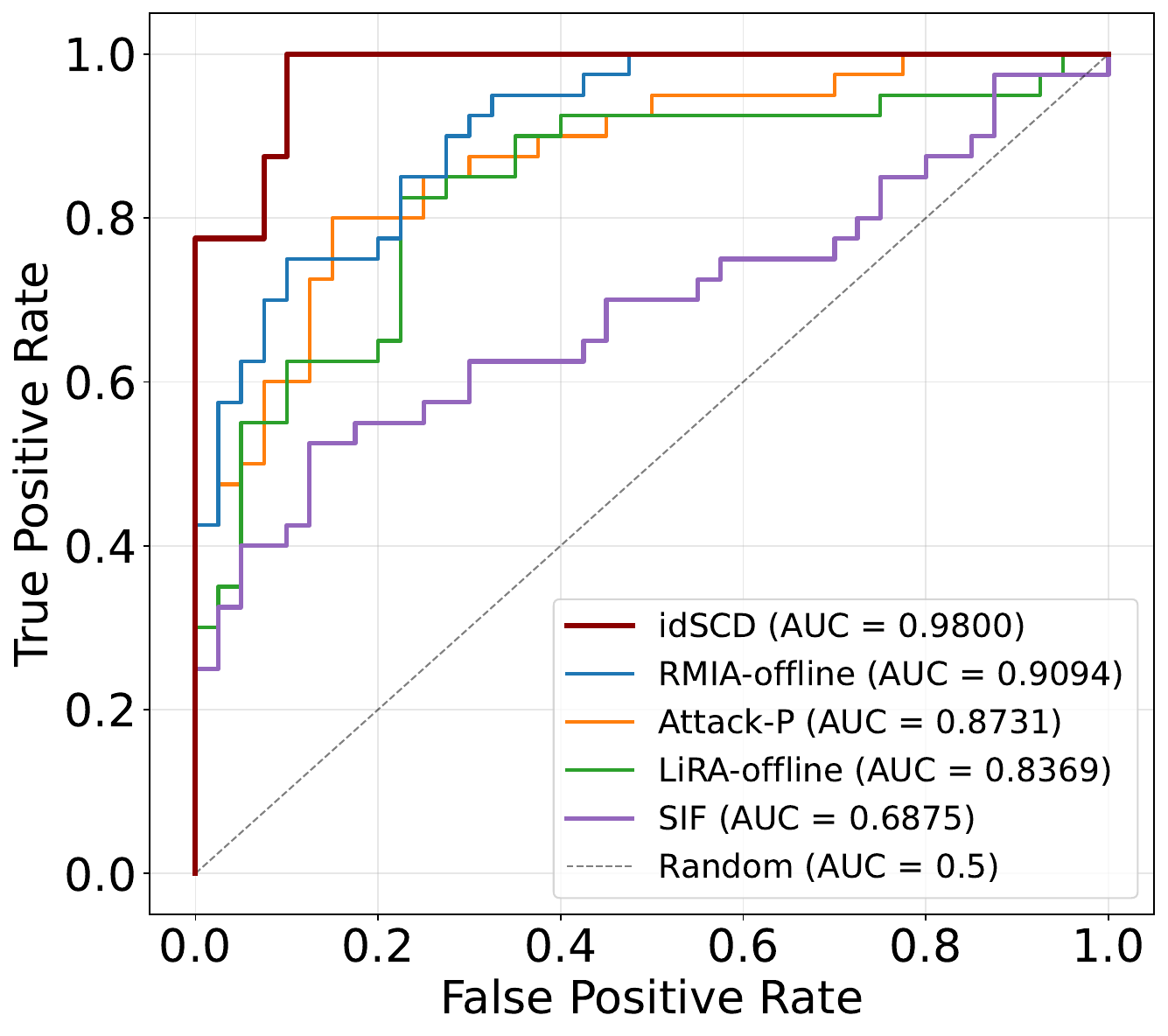}
        \caption{Emotion group.}
        \label{fig:sota_emotion}
    \end{subfigure}
    \begin{subfigure}[t]{0.24\linewidth}
        \centering
        \includegraphics[width=\linewidth]{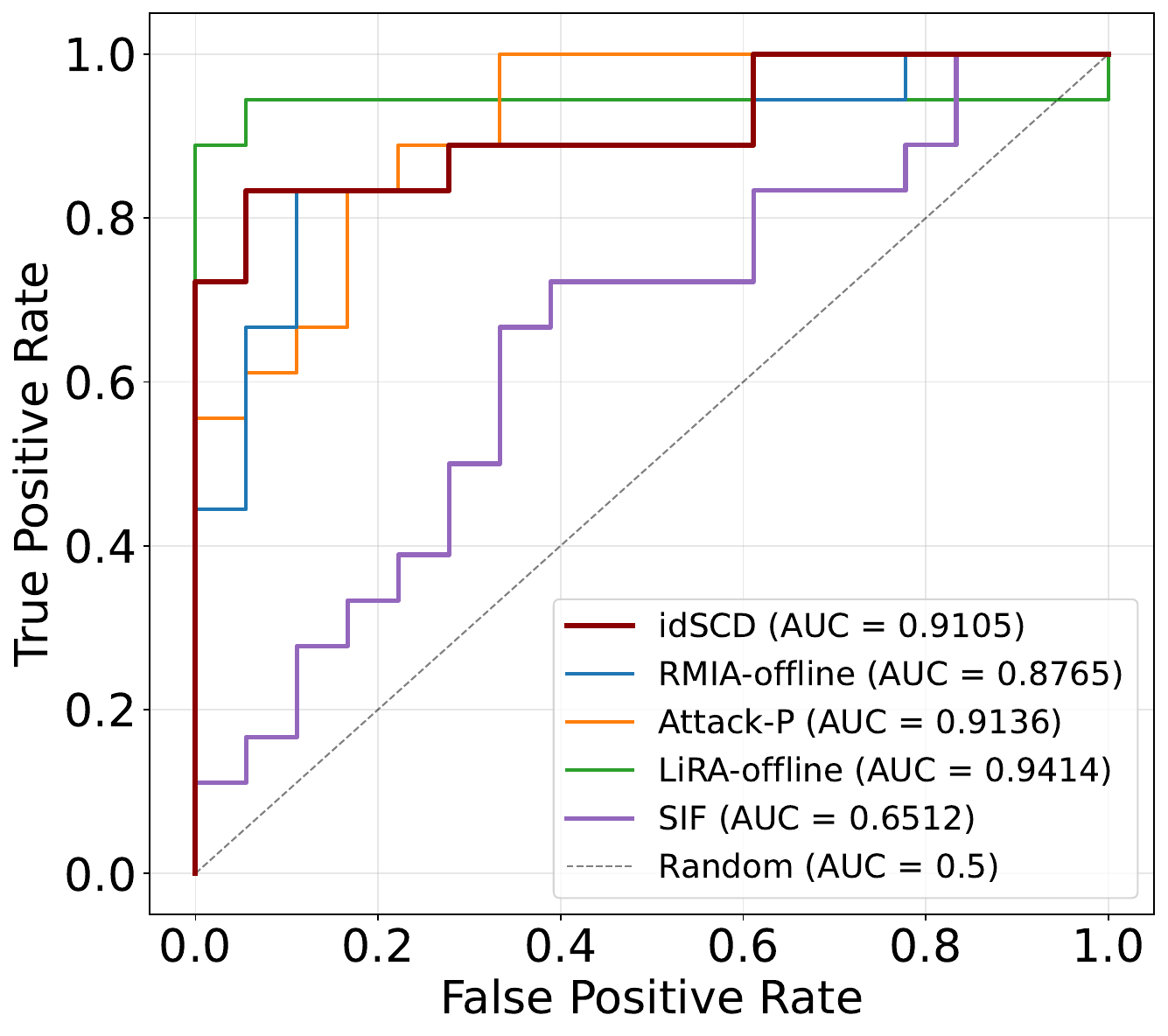}
        \caption{NLI datasets group.}
        \label{fig:sota_nli}
    \end{subfigure}
    \begin{subfigure}[t]{0.24\linewidth}
        \centering
        \includegraphics[width=\linewidth]{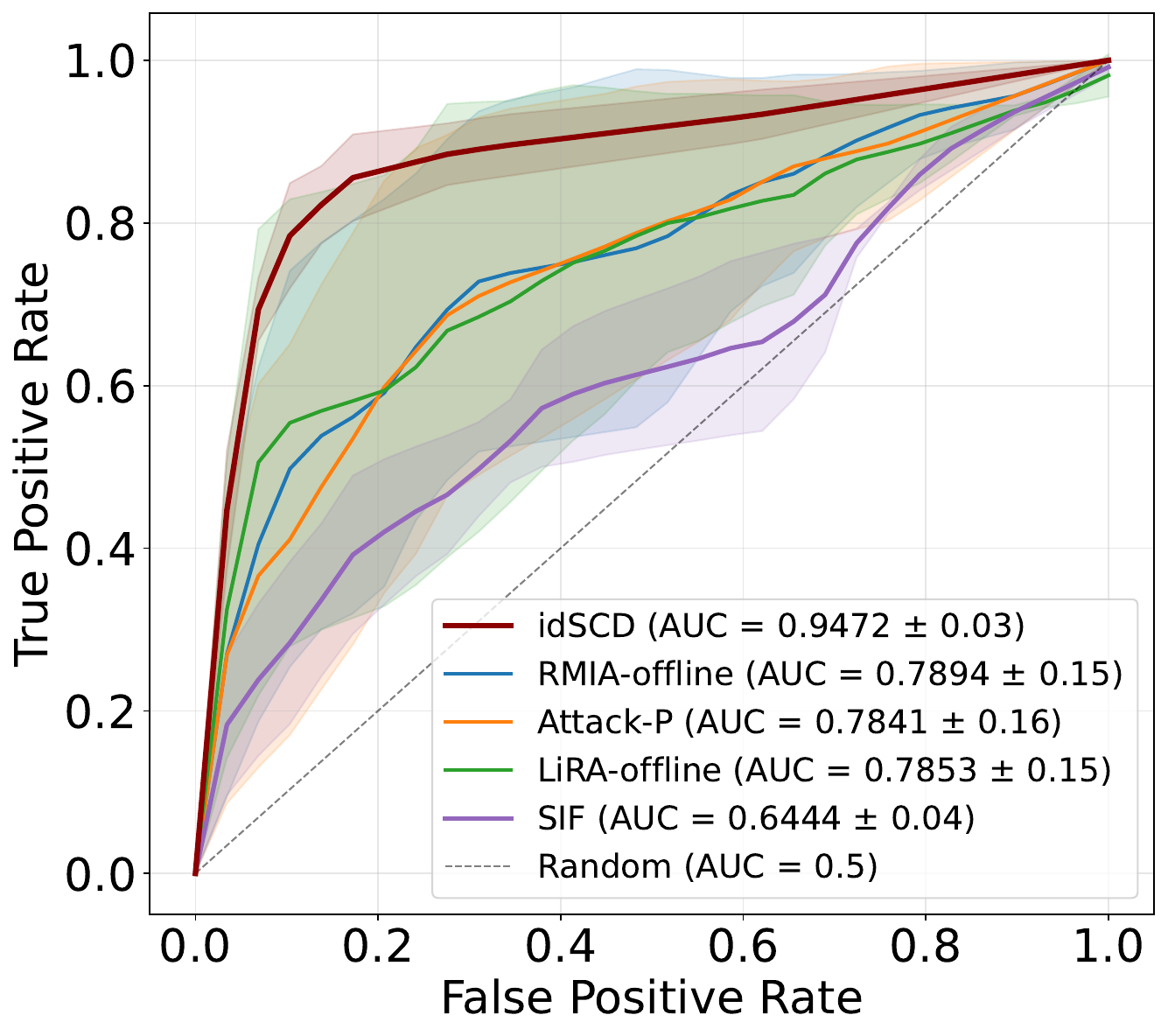}
        \caption{Mean across groups.}
        \label{fig:sota_mean_std}
    \end{subfigure}
    \caption{SOTA comparison of ROC curves for dataset membership inference. (a-c) show performance on each dataset group, while (d) summarizes the mean and standard deviation (std) across groups. We compare against black-box baselines RMIA, Attack-P, and LiRA, and white-box SIF baseline. We achieve the top performance on the medical and emotion groups, are competitive on NLI, and obtain the highest average performance with lowest std. In contrast, the baselines fluctuate more across groups, suggesting they are more sensitive to non-semantic dataset-specific artifacts.}
    \label{fig:sota_comparison}
\end{figure}

\noindent\textbf{Results.} Fig.~\ref{fig:sota_comparison} shows that our SCD-based classifier achieves the strongest performance on the medical and emotion groups, and remains competitive on NLI. The largest gains appear on the medical group, where \(id_{\mathrm{SCD}}\) improves over all baselines by more than \(60\%\) (see the discussion below for why this is the case). On emotion, \(id_{\mathrm{SCD}}\) also achieves the best performance, improving over the strongest baseline by about \(8\%\) and reaching a ROC-AUC of \(0.980\). On NLI, \(id_{\mathrm{SCD}}\) remains competitive, but is no longer the top method: it is about \(3\%\) below the best baseline. As discussed below, this gap is explained by the much smaller shared semantic support in the NLI setup, a condition that can be detected before applying our method and used to assess whether SCD-based membership inference is suitable for a given dataset group. Aggregated across groups, \(id_{\mathrm{SCD}}\) achieves the highest mean ROC-AUC and the lowest std, indicating a stronger and more stable membership signal. This suggests that SCDs are most effective when datasets induce distinct semantic correlation patterns, while baselines are more sensitive to non-semantic dataset artifacts.

\noindent\textbf{Why medical benefits from SCDs.}
The strongest improvement (+60\% in ROC-AUC) appears on the medical dataset group, and this is expected from the construction of the benchmark. Unlike NLI and emotion, the medical group is built by splitting each class in PubMed 200k RCT into five topic clusters using high-level semantic features. This creates datasets that differ substantially in their semantic content while preserving the same rhetorical-role label space. Such a construction directly favors methods that use internal semantic correlations, because the resulting splits induce more separated feature-class semantic association patterns. In contrast, output-based attacks may not fully capture these high-level semantic differences if the model's predictive behavior remains similar across clusters. The medical result therefore highlights the main advantage of SCDs: when datasets differ semantically, their internal correlation fingerprints can be more discriminative than output behavior.

\noindent\textbf{Limited SCD suitability on NLI.}
The smaller margin on NLI is also informative. Our method compares semantic correlations over a keyword vocabulary shared across datasets, so the effective decision signal depends on how much semantic overlap is exposed through shared keywords between the target dataset and the inspected model's descriptor. In the NLI group, the keyword overlap between two datasets is only 15\%, while in the other groups is 20\%. This leaves the SCD classifier with less shared semantic evidence and explains why the SCD signal is less dominant on NLI: the method is designed to exploit correlations between shared keywords and classes, and its discriminative power decreases when the common keyword support is sparse. See App.~\ref{app:keyword-overlap} for details.
 
\noindent\textbf{Takeaway.}
These results suggest that semantic correlation structure can substantially change how dataset-level membership is detected: when datasets induce distinct semantic fingerprints, gains over output-based and influence-based baselines can be large. Across our three evaluation groups, SCDs are strongest when datasets are semantically separable and share enough keyword support, while the baselines oscillate much more, suggesting higher sensitivity to non-semantic dataset-specific artifacts. Importantly, shared-keyword support can be checked a priori, providing a simple diagnostic for whether SCD-based inference is sufficiently semantically informed for a given dataset group.

\section{Limitations}
\label{sec:limitations}

\noindent\textbf{White-box access.}
Our strongest instantiation of the method uses internal model information,
including class weights and embedding-space associations. This is a limitation
for fully closed models, where such quantities are not directly available.
However, this assumption matches many realistic auditing settings, including
internal model audits, open-weight models, model handoffs, and regulated
provenance checks where the auditor can inspect model parameters. Moreover, the
framework is not intrinsically tied to this exact signal: the correlation
descriptor can be instantiated with alternative scoring functions, including
output-based probes or descriptors extracted from distilled surrogate models.
These variants may provide weaker evidence, but they offer a path toward
applying the same dataset-level tracing principle beyond strict white-box
access.
 
\noindent\textbf{Dataset identity versus dataset similarity.}
Our method operates at the level of dataset descriptors, not individual training
examples. A strong match should therefore be interpreted as evidence that the
model contains a dataset-level correlation pattern associated with \(D_i\), not
as a sample-level proof that every example in \(D_i\) was used during training.
This distinction matters when datasets share domain, style, label semantics, or
annotation artifacts: a high score may reflect genuine use of \(D_i\), but it
may also arise from a closely related dataset with a similar correlation
structure. The leave-one-dataset-out analysis tests whether the recovered
contribution is better explained by \(D_i\) than by other datasets,
which reduces this ambiguity within the evaluated dataset groups. Still,
absolute dataset identity is difficult when two datasets are near
duplicates or intentionally constructed to share the same artifacts.
 
\noindent\textbf{Dependence on shared keyword support.}
SCD-based inference relies on comparing semantic correlations over keywords that are shared across datasets. When the overlap between dataset-specific keyword vocabularies is sparse, the descriptor comparison has less semantic evidence to exploit, and the membership signal can become weaker. This limitation is partly diagnosable before running the method: low shared-keyword overlap indicates that the dataset group may be less suitable for SCD-based membership inference. In our experiments, this effect is visible in the NLI group, where lower keyword overlap reduces the discriminative power of the SCD classifier.
 
\noindent\textbf{Class alignment (evaluation framework limitation).}
Our evaluation requires datasets within a group to be mapped to a shared set of
canonical labels before fingerprinting. This is primarily a constraint of the
current test framework, rather than a fundamental requirement that all datasets
share identical labels. In practice, many scenarios already
involve groups of datasets for the same task, such as NLI, emotion
classification, or medical text classification, where such mappings are
meaningful. We evaluate the method across three dataset groups to avoid relying
on a single label taxonomy. But the quality of canonicalization
matters: ambiguous, hierarchical, or partially overlapping labels can
blur correlation descriptors and make dataset-specific patterns harder to
separate.
 
\noindent\textbf{Scalability of controlled validation (evaluation framework limitation).}
The leave-one-dataset-out protocol used in our controlled sensitivity analysis
requires training one model per removed dataset, which becomes expensive as the
number of candidate datasets grows. This cost is a limitation of the validation
protocol, not of the final membership score itself. The controlled setting is
useful because it isolates the contribution of each dataset and tests whether
correlation descriptors respond to dataset-specific changes. For larger
collections, this ablation-based analysis could be replaced or approximated by
more efficient strategies, such as subset ablations, influence-style
approximations, low-rank descriptor updates, or surrogate models. Importantly,
our more realistic membership scoring setting does not require training a full
leave-one-dataset-out suite.

\noindent\textbf{Semantic similarity vs. exact dataset membership} Our method detects dataset-specific semantic correlation patterns rather than establishing training dataset identity with certainty. Leave-one-dataset-out diagnostics favor the true member within the evaluated groups, but semantically similar or near-duplicate non-members may also score highly. Without access to the original training data, dataset-level membership inference relies on statistical and semantic traces rather than exact identity.

\section{Conclusion}
\label{sec:conclusion}

We show that a dataset can be recognized from the semantic correlations it induces during training. Our work reframes dataset-level membership inference as a white-box semantic fingerprinting problem: instead of relying on output behavior, we trace datasets through the internal correlation structure learned by the model. We instantiate this idea with Semantic Correlation Descriptors (SCDs), show in a controlled leave-one-dataset-out diagnostic that they recover dataset-specific changes with perfect separation, and introduce a practical SCD-based membership score that only requires the inspected model's SCD and the target dataset's standalone SCD. Across NLI, emotion classification, and medical text classification, we evaluate SCDs under different levels of semantic separation and keyword support. On average, the resulting \(id_{\mathrm{SCD}}\) classifier achieves the highest performance and the lowest std among recent strong baselines, including RMIA, Attack-P, LiRA, and SIF. The largest gains appear when dataset groups expose distinct semantic particularities, reaching up to \(+60\%\) ROC-AUC. These results establish internal semantic correlations as a strong and stable signal for dataset provenance, complementary to behavioral and influence-based membership inference.

\begin{ack}
This work is supported in part by projects “Romanian
Hub for Artificial Intelligence - HRIA”, Smart Growth, Digitization and Financial Instruments Program, 2021-2027 (MySMIS no. 334906) and "European Lighthouse of AI for Sustainability - ELIAS", Horizon Europe program (Grant No. 101120237).

\end{ack}

\newpage
\bibliography{main}
\bibliographystyle{plainnat}

\newpage
\appendix

\section{Appendix}

\subsection{Broader Impact}
\label{sec:broader_impact}
This work supports training-data auditing by providing a white-box signal for dataset attribution and dataset-level membership inference. Potential positive uses include dataset provenance analysis, benchmark contamination detection, licensing audits, and accountability for models trained on mixed data sources. The method is also dual-use. The same signal could be used to infer confidential or commercially sensitive information about a model's training data. For this reason, the results should be interpreted as statistical evidence rather than definitive proof of dataset use, especially in legal, privacy-sensitive, or compliance settings.

\subsection{Dataset configurations and training regimes}
\label{app:dataset_configurations}

Tab.~\ref{apx:tab:dataset_sizes} reports the dataset configurations used in our experiments. For each dataset group, we construct three types of training sets: individual datasets, leave-one-dataset-out mixtures, and the full combined corpus. The individual configurations provide standalone dataset fingerprints, the leave-one-dataset-out configurations support the controlled sensitivity analysis, and the full corpus configurations are used for dataset-level membership evaluation.

Overall, this yields 27 training configurations across the three experimental groups: 9 for emotion classification, 7 for native language identification, and 11 for medical text classification. The table also reports the number of training rows in each configuration, making explicit the scale of each model training regime used in the main experiments.

\begin{table}[t]
    \caption{Dataset configurations and sample sizes across the three experimental groups. For each group, we construct individual, leave-one-dataset-out, and full-corpus training sets, yielding 27 training configurations in total.}
    \centering
    \small
    \begin{tabular}{@{}llc@{}}
        \toprule
        \textbf{Dataset group} & \textbf{Training configuration} & \textbf{Sample size (rows)} \\
        \midrule
        \textbf{Emotion} & EmoLit (Individual) & 200,727 \\
        & EmotionDataset20 (Individual) & 79,595 \\
        & GoEmotions (Individual) & 53,993 \\
        & XED \& SMED (Individual) & 92,978 \\
        \cmidrule{2-3}
        & Without EmoLit (Leave-One-Out) & 226,563 \\
        & Without EmotionDataset20 (Leave-One-Out) & 347,695 \\
        & Without GoEmotions (Leave-One-Out) & 373,300 \\
        & Without XED \& SMED (Leave-One-Out) & 334,315 \\
        \cmidrule{2-3}
        & \textbf{Full Combined Corpus} & \textbf{427,290} \\
        \midrule
        \textbf{NLI} & iTalki (Individual) & 45,204 \\
        & Reddit-L2 (Individual) & 45,201 \\
        & EFCAMDAT (Individual) & 45,200 \\
        \cmidrule{2-3}
        & Without iTalki (Leave-One-Out) & 90,401 \\
        & Without Reddit-L2 (Leave-One-Out) & 90,404 \\
        & Without EFCAMDAT (Leave-One-Out) & 90,405 \\
        \cmidrule{2-3}
        & \textbf{Full Balanced Corpus} & \textbf{135,605} \\
        \midrule
        \textbf{Medical} & Cluster 0 (Individual) & 21,100 \\
        & Cluster 1 (Individual) & 31,791 \\
        & Cluster 2 (Individual) & 33,251 \\
        & Cluster 3 (Individual) & 160,516 \\
        & Cluster 4 (Individual) & 25,776 \\
        \cmidrule{2-3}
        & All Except Cluster 0 (Leave-One-Out) & 251,334 \\
        & All Except Cluster 1 (Leave-One-Out) & 240,643 \\
        & All Except Cluster 2 (Leave-One-Out) & 239,183 \\
        & All Except Cluster 3 (Leave-One-Out) & 111,918 \\
        & All Except Cluster 4 (Leave-One-Out) & 246,658 \\
        \cmidrule{2-3}
        & \textbf{Full Combined Corpus (12\% Subsample)} & \textbf{272,434} \\
        \bottomrule
    \end{tabular}
    \label{apx:tab:dataset_sizes}
\end{table}

\subsection{Per-class keyword overlap between single-source datasets}
\label{app:keyword-overlap}

To better understand why our membership inference attack performs differently across dataset groups, we measure how much the discriminative vocabularies of individual source datasets overlap. A low overlap indicates that each source contributes a largely unique set of class-discriminative features, leaving less shared signal for the fingerprint-based detector to exploit when comparing models trained on different subsets.

\paragraph{Procedure.}
For each dataset group (NLI, Emotion, Medical), we extract the top 5{,}000 keywords independently from each single-source training set using YAKE (with $n$-gram size 3 and deduplication threshold 0.9). For each single-source configuration $i$, we then identify which of its keywords appear in the concatenated (lowercased) texts of each class $c$, yielding a set of per-class keyword occurrences. We denote the number of keywords from configuration $i$ that appear in class $c$ of its own dataset as part of this set.

To quantify the overlap between two single-source configurations $i$ and $j$, we count, for each class $c$, how many keywords are found in the texts of class $c$ in \emph{both} configurations, and sum this count across all classes. This gives the raw overlap value reported in the tables below as the off-diagonal entries. The diagonal entries represent the total number of per-class keyword occurrences within each configuration's own data (i.e., self-overlap).

To obtain a normalized measure, we divide each pairwise overlap value by the average of the two corresponding diagonal (self-overlap) values. This yields a ratio between 0 and 1, where 1 would indicate identical per-class keyword coverage and 0 would indicate completely disjoint vocabularies. We then report the mean of all off-diagonal normalized values as the average pairwise overlap for each dataset group.

\paragraph{Results.}

Tab.~\ref{tab:overlap-nli} shows the normalized pairwise overlap matrix for the NLI group, Tab.~\ref{tab:overlap-emotion} for the Emotion group, and Tab.~\ref{tab:overlap-medical} for the Medical group. Tab.~\ref{tab:overlap-summary} summarizes the average pairwise overlap across groups alongside the $id_{\mathrm{SCD}}$ ROC-AUC.

\begin{table}[h]
\centering
\caption{Normalized per-class keyword overlap between single-source datasets (NLI).}
\label{tab:overlap-nli}
\begin{tabular}{lccc}
\toprule
 & reddit\_l2 & efcamdat & italki \\
\midrule
reddit\_l2 & 1.00 & 0.14 & 0.14 \\
efcamdat   & 0.14 & 1.00 & 0.17 \\
italki     & 0.14 & 0.17 & 1.00 \\
\bottomrule
\end{tabular}
\end{table}

\begin{table}[h]
\centering
\caption{Normalized per-class keyword overlap between single-source datasets (Emotion).}
\label{tab:overlap-emotion}
\begin{tabular}{lcccc}
\toprule
 & EmoLit & Emotion20 & GoEmotions & XED\_SMED \\
\midrule
EmoLit      & 1.00 & 0.12 & 0.24 & 0.17 \\
Emotion20   & 0.12 & 1.00 & 0.20 & 0.20 \\
GoEmotions  & 0.24 & 0.20 & 1.00 & 0.25 \\
XED\_SMED   & 0.17 & 0.20 & 0.25 & 1.00 \\
\bottomrule
\end{tabular}
\end{table}

\begin{table}[h]
\centering
\caption{Normalized per-class keyword overlap between single-source datasets (Medical).}
\label{tab:overlap-medical}
\begin{tabular}{lccccc}
\toprule
 & split\_0 & split\_1 & split\_2 & split\_3 & split\_4 \\
\midrule
split\_0 & 1.00 & 0.17 & 0.12 & 0.16 & 0.13 \\
split\_1 & 0.17 & 1.00 & 0.25 & 0.27 & 0.25 \\
split\_2 & 0.12 & 0.25 & 1.00 & 0.18 & 0.19 \\
split\_3 & 0.16 & 0.27 & 0.18 & 1.00 & 0.27 \\
split\_4 & 0.13 & 0.25 & 0.19 & 0.27 & 1.00 \\
\bottomrule
\end{tabular}
\end{table}

\begin{table}[h]
\centering
\caption{Average pairwise normalized keyword overlap (single-source pairs only) and $id_{\mathrm{SCD}}$ membership inference ROC-AUC for each dataset group.}
\label{tab:overlap-summary}
\begin{tabular}{lccc}
\toprule
Dataset Group & Single Sources & Avg.\ Pairwise Overlap (\%) & $id_{\mathrm{SCD}}$ ROC-AUC \\
\midrule
NLI     & 3 & 14.96 & 0.9105 \\
Emotion & 4 & 19.62 & 0.9800 \\
Medical & 5 & 20.01 & 0.9511 \\
\bottomrule
\end{tabular}
\end{table}

\paragraph{Implications.}
The NLI group exhibits the lowest average pairwise overlap (14.96\%), meaning that its constituent sources (reddit\_l2, efcamdat, italki) share only about 15\% of their class-discriminative vocabulary. By contrast, the Emotion and Medical groups share approximately 20\% of theirs. This aligns with the lower $id_{\mathrm{SCD}}$ ROC-AUC observed for NLI: when single-source datasets have less overlapping discriminative features, the aligned keyword matrices used by the detector contain more zero-filled (uninformative) positions, which dilutes the similarity signal that distinguishes member from non-member configurations. In other words, less lexical overlap between sources means less information is available to inform the dataset fingerprint, making the membership inference task harder.

Importantly, this overlap can be measured \emph{a priori}, before training any models or running the attack, using only the raw text data and an unsupervised keyword extraction method. This suggests that keyword overlap between constituent sources could serve as a practical diagnostic for anticipating the difficulty of fingerprint-based dataset membership inference in multi-source settings.


\clearpage

\subsection{Multi-seed experiments}
\label{app:multi-seed-experiments}
To test the stability of our results, we repeat the main experiment over 5 independent random seeds (0, 1, 2, 3, 42). The seed controls model initialization, data shuffling,
and the sampling of member/non-member datasets. All other hyperparameters are
kept identical to those in Section~\ref{sec:experiments}.
Table~\ref{tab:multi_seed} reports the mean and standard deviation across seeds. 

The results are consistent across seeds, confirming that the conclusions of Section~\ref{sec:experiments} do not depend on
a particular random seed.

\begin{table}[h]
\centering
\caption{ Results over 5 random seeds (mean $\pm$ std).}
\label{tab:multi_seed}
\begin{tabular}{lccc}
\toprule
Dataset group & ROC-AUC \\
\midrule
Emotion      & 0.990 $\pm$ 0.006 \\
NLI          & 0.914 $\pm$ 0.007  \\
Medical          & 0.955 $\pm$ 0.021  \\
\bottomrule
\end{tabular}
\end{table}

\subsection{Quantitative Descriptor Size Ablations}

We vary the number of extracted keywords per dataset ($k$), which changs the size of the shared keyword vocabulary ($|U_k|$) and consequently the descriptor dimensionality ($|U_k| \times |Y|$). Across all three dataset groups, increasing the descriptor dimensionality consistently improves ROC-AUC before largely saturating, as reported in Table \ref{tab:descriptor_dim}

\begin{table}[h]

\centering
\caption{ Descriptor dimensionality ablation results}
\label{tab:descriptor_dim}
\begin{tabular}{lccc}
\toprule
Dataset group & k & Descriptor dimension & ROC-AUC\\
\midrule
Emotion      & 10 & 396 & 0.740 \\
Emotion      & 50 & 1818 & 0.913 \\
Emotion      & 100 & 3528 & 0.968 \\
Emotion      & 500 & 17757 & 0.965 \\
Emotion      & 1000 & 35136 & 0.977 \\
\textbf{Emotion}      & \textbf{5000} & \textbf{169425} & \textbf{0.980} \\
\midrule
NLI      & 10 & 190 & 0.824 \\
NLI      & 50 & 915 & 0.818 \\
NLI      & 100 & 1835 & 0.855 \\
NLI      & 500 & 8650 & 0.880 \\
NLI      & 1000 & 16995 & 0.886 \\
\textbf{NLI}      & \textbf{5000} & \textbf{81550} & \textbf{0.910} \\

\midrule
Medical      & 10 & 265 & 0.737 \\
Medical      & 50 & 1145 & 0.856 \\
Medical      & 100 & 2255 & 0.878 \\
Medical      & 500 & 10935 & 0.877 \\
Medical      & 1000 & 21545 & 0.918 \\
\textbf{Medical}      & \textbf{5000} & \textbf{104630} & \textbf{0.951} \\
\bottomrule
\end{tabular}
\end{table}

\subsection{Independently seeded descriptor controls}
To evaluate the fingerprint robustness cross-seed, we compute reference descriptors from independently seeded single-dataset models with multiple independent random seeds while keeping the full and leave-one-dataset-out models fixed. We then recomputed the controlled diagnostic by replacing each reference descriptor with the descriptor obtained from the independently seeded model and regenerated the corresponding matrices. 
We use the descriptors obtained with seed \(42\) as a fixed reference and compare them with descriptors learned independently using seeds \(s \in \{0,1,2,3\}\). The resulting correlation structure is highly consistent across the four comparison seeds. Both the sign and magnitude of the correlations vary only slightly from one seed to another. In particular, the diagonal correlations remain positive for all descriptors, while the off-diagonal relationships exhibit nearly identical patterns across seeds. This suggests that, although the descriptors obtained from different random initializations are not identical, their relationship to the seed-\(42\) reference representation is reproducible.
 This shows that the alignment between $\Delta_{i}$
 and 
 $d^{(i)}_{SCD}$ is robust to independent training initializations and reflects stable dataset-specific semantic fingerprints rather than a seed-specific construction artifact. 

 \begin{figure*}[t]
    \centering
    \begin{subfigure}[c]{0.32\textwidth}
        \centering
        \includegraphics[width=\linewidth]{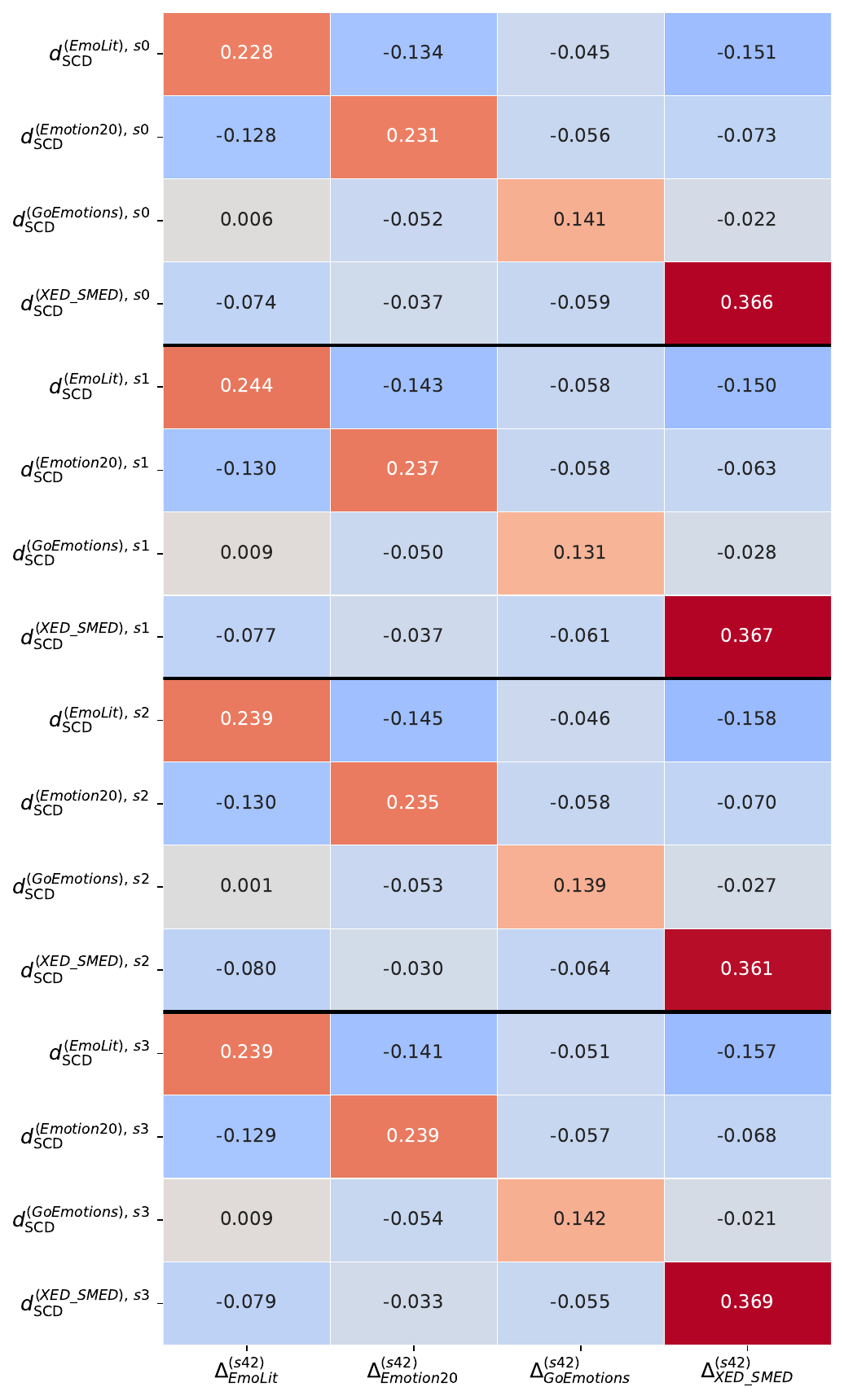}
        \caption{Emotion}
        \label{fig:crossseed_emotions}
    \end{subfigure}
    \hfill
    \begin{subfigure}[c]{0.32\textwidth}
        \centering
        \includegraphics[width=\linewidth]{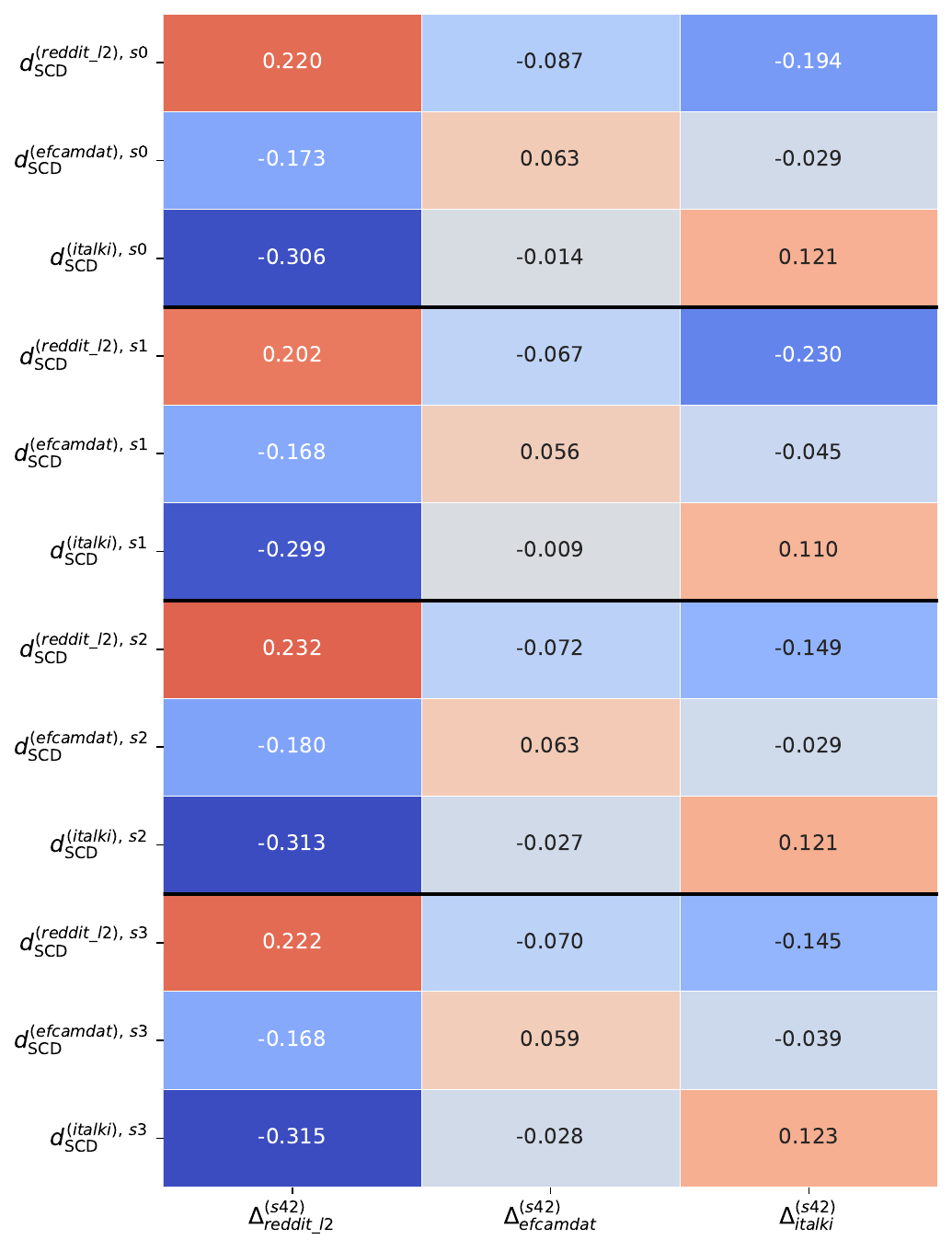}
        \caption{NLI}
        \label{fig:crossseed_nli}
    \end{subfigure}
    \hfill
    \begin{subfigure}[c]{0.32\textwidth}
        \centering
        \includegraphics[width=\linewidth]{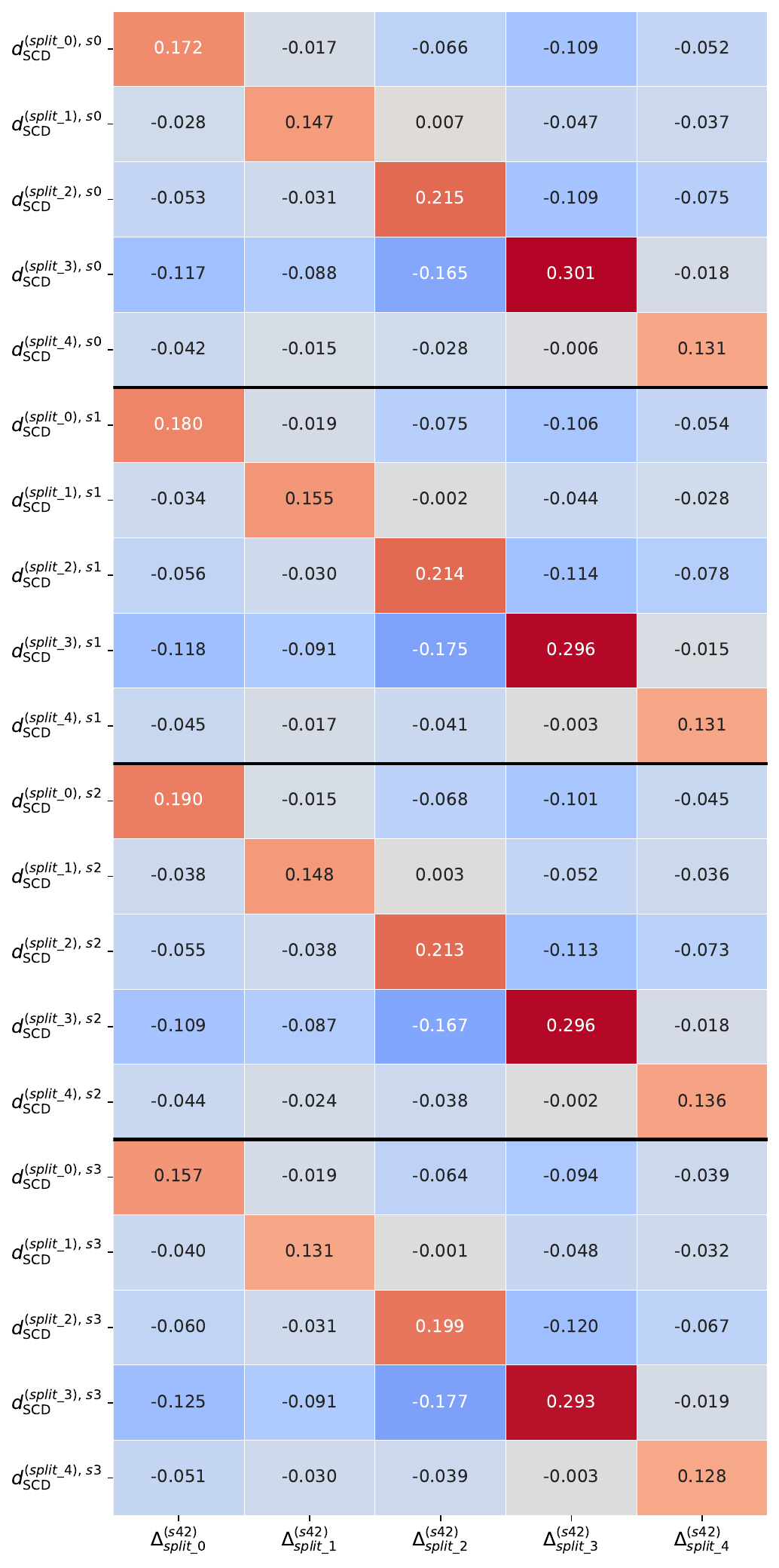}
        \caption{Medical}
        \label{fig:crossseed_medical}
    \end{subfigure}
    \caption{Cross-seed descriptor reproducibility relative to seed 42. Each cell reports the correlation between a descriptor learned with a comparison seed \(s \in \{0,1,2,3\}\) (rows) and a descriptor learned with the reference seed \(42\) (columns). Diagonal entries compare corresponding descriptors across seeds, while off-diagonal entries capture cross-descriptor relationships. Horizontal lines separate the four comparison seeds.}
    \label{fig:crossseed}
\end{figure*}

 \subsection{ IID and Non-IID Partitions}
We test whether the SCD fingerprint depends on the specific training examples or on
the data distribution. To do so, we split each dataset $D_i$ into two disjoint parts
and train a model on each part. Training uses the same procedure, hyperparameters and
seed as the single-dataset models. We use two partitioning schemes:

\begin{itemize}
    \item \textbf{IID:} a random 50/50 split. Both halves preserve the class
    distribution of $D_i$.
    \item \textbf{Non-IID:} each class $c$ is sampled at its own rate $r_c$, spaced
    linearly from $0.02$ to $0.98$ across the nine classes. A fraction $r_c$ of class
    $c$ goes to the first part and $1 - r_c$ to the second. The two parts therefore have
    strongly skewed, roughly mirrored class distributions; for example, \emph{Trust}
    makes up under $1\%$ of the second part. Because the rates average $0.5$, the two
    parts remain similar in size.
\end{itemize}

To keep all descriptors in the same space, each part model reuses its parent dataset's
keyword list. We then correlate each part model's descriptor with the difference
descriptors $\Delta_j$ computed from the original full-data models.

\begin{figure}[h]
    \centering
    \includegraphics[width=0.6\linewidth]{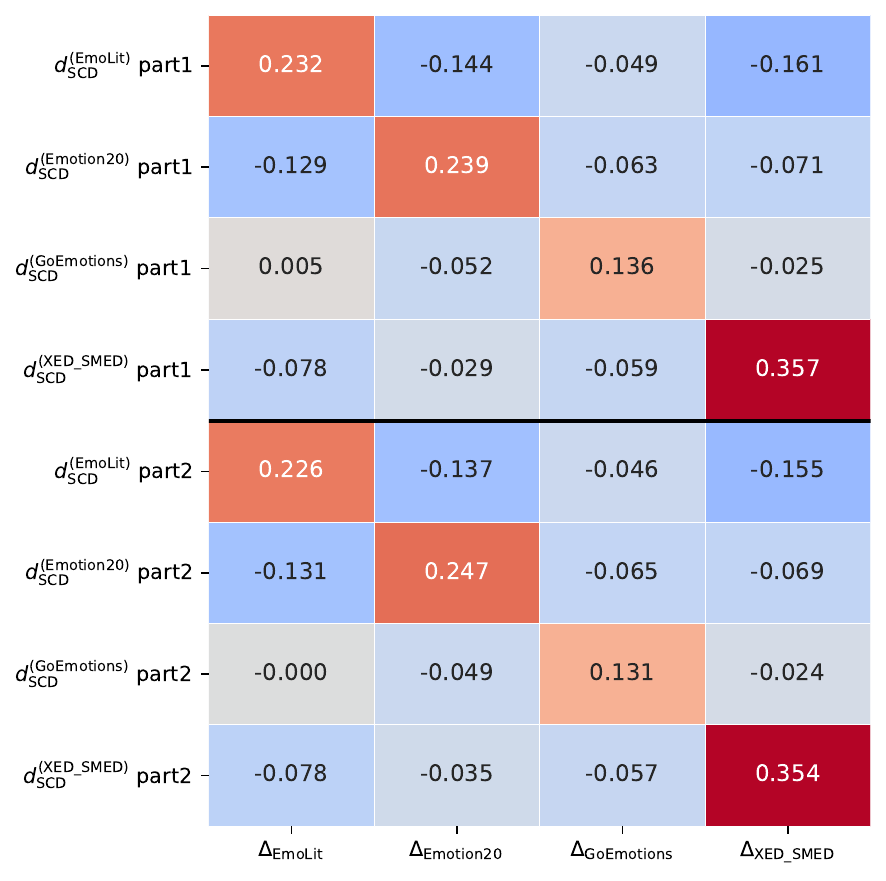}
    \caption{IID split. Correlation between $\Delta_i$ and the descriptors
    of models trained on each random half.}
    \label{fig:subset_iid}
\end{figure}

\begin{figure}[h]
    \centering
    \includegraphics[width=0.6\linewidth]{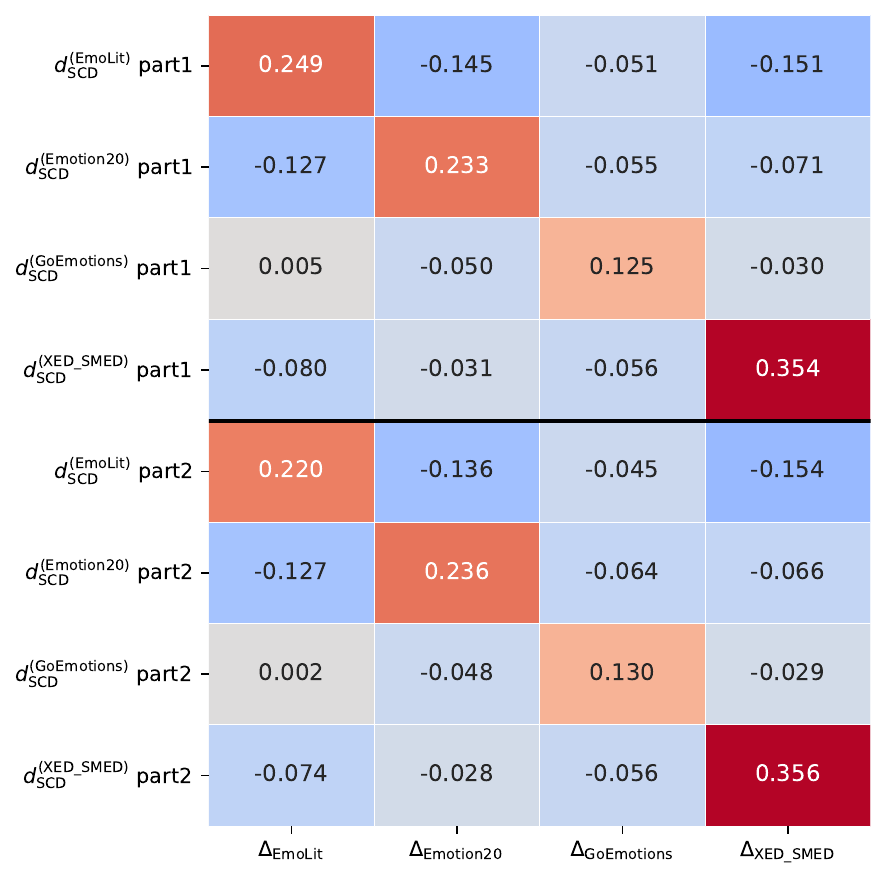}
    \caption{Non-IID split. Correlation between $\Delta_i$ and the descriptors of models trained on the two class-skewed parts of each dataset.}
    \label{fig:subset_noniid}
\end{figure}

Figures~\ref{fig:subset_iid} and~\ref{fig:subset_noniid} show the results. Under both
schemes, every part remains diagonally dominant for all four datasets, and the
off-diagonal structure is essentially unchanged. The fingerprint therefore behaves mainly as a
property of the data distribution rather than of individual examples. The experiment shows that a subset reproduces the dataset's signature:
it does not compare examples a model was trained on against examples it never saw.

\end{document}